\documentclass[11pt,letterpaper]{article}
\usepackage[top=1in,bottom=1in,left=1in,right=1in]{geometry}
\usepackage{natbib}      
\usepackage{palatino}
\bibpunct{(}{)}{;}{a}{,}{,}

\usepackage{chngpage}
\usepackage{stmaryrd}
\usepackage{amssymb}
\usepackage{amsmath}
\usepackage{graphicx}
\usepackage{lscape}
\usepackage{subfigure}
\usepackage[usenames,dvipsnames]{color}
\definecolor{myblue}{rgb}{0,0.1,0.6}
\definecolor{mygreen}{rgb}{0,0.3,0.1}
\usepackage[colorlinks=true,linkcolor=black,citecolor=mygreen,urlcolor=myblue]{hyperref}

\usepackage{multirow}

\newcommand{\ignore}[1]{}

\newenvironment{itemizesquish}{\begin{list}{\labelitemi}{\setlength{\itemsep}{0em}\setlength{\labelwidth}{0.5em}\setlength{\leftmargin}{\labelwidth}\addtolength{\leftmargin}{\labelsep}}}{\end{list}}
\newenvironment{examples}{\begin{itemizesquish}}{\end{itemizesquish}}

\title{
  ARKref: a rule-based coreference resolution system
}

\author{
  Brendan O'Connor$^\ast$ and Michael Heilman$^\dagger$
  \\ \\
  $\ast$: Carnegie Mellon University 
  \\
  $\dagger$: Educational Testing Service 
}

\begin{document}
\maketitle

\section{Introduction}

ARKref is a tool for noun phrase coreference that 
is based on the system 
described by \citet{haghighi-09} (which was never publicly released).
It was originally written in 2009.  At the time of writing, 
the last released version was in March 2011.
This document describes that version, which is open-source and publicly available at \url{http://www.ark.cs.cmu.edu/ARKref}.\footnote{\citet{Heilman2011Thesis} previously provided a less detailed description of the ARKref system.}

ARKref is a deterministic, rule-based system
that uses syntactic information from a constituent parser, 
and semantic information from an entity recognition component,
to constrain the set of possible mention candidates (i.e., noun phrases) that could be 
antecedents for a given mention.
It encodes syntactic constraints such as the fact that 
the noun phrases in predicative nominative constructions corefer 
(e.g., \emph{John was the teacher.}), as well as semantic
constraints such as the fact that \emph{he} cannot corefer
with a noun labeled as a location.
After filtering candidates with these constraints,
it selects as the antecedent the 
candidate noun phrase
with the shortest (cross-sentence) tree distance
from the target.  Antecedent decisions are aggregated with a transitive closure
to create the final entity graph.

ARKref belongs to a family of rule-based coreference systems that use rich
syntactic and semantic information to make antecedent selection decisions.
Besides \citeauthor{haghighi-09},
current work in this vein includes \cite{Lee2013Coref}, which was one of the
best performing systems in a recent CoNLL shared task.

The following example provides an illustration of ARKref's output, 
in which brackets denote the extent of noun phrases and
indices denote the entity to which each noun phrase refers. 
This example emphasizes the syntactic selection criteria:
\begin{examples}
\item $[$John$]_1$ bought $[$himself$]_1$ $[$a book$]_2$ .
$[$Fred$]_3$ found out that $[$John$]_1$ had also bought $[$himself$]_1$ $[$a computer$]_4$ .
Next, $[$he$]_3$ found out that $[$John$]_1$ had not bought $[$him$]_3$ $[$anything$]_5$ . \label{ex:arkref}
\end{examples}

\section{System description}
The system can be described either in terms of differences from previous work, or by itself.

\subsection{Comparison to \citet{haghighi-09}}

There are a few key differences between ARKref
and the approach of \citet{haghighi-09}:

\begin{itemizesquish}
\item ARKref does not include the bootstrapped lexical semantic 
      compatibility subsystem (``\texttt{SEM-COMPAT}'').
\item ARKref uses the supersense tagger originally described by \citet{ciaramita-altun-06} and reimplemented by \citet{Heilman2011Thesis} (\S 3.1.3)
      rather than a named entity recognizer to match entity types, allowing
      it to incorporate additional information about common nouns.
\item ARKref encodes a few additional syntactic constraints on coreference:
      Objects cannot refer to subjects unless
      they are reflexive (e.g., in \emph{John called him}, the pronoun \emph{him} cannot refer to \emph{John});
      and subjects cannot refer to sentence-initial adjuncts
      (e.g., in \emph{To call John, he picked up the phone}, the subject \emph{he} cannot refer to \emph{John}).  
      Binding theory (\citealp{chomsky-81}; \citealp[Chapter 4]{carnie-06}) provides general explanations that explain these constraints as well as others not encoded in the system.
\item ARKref uses male and female name lists from the 1990 U.S. Census\footnote{The census name lists can be downloaded at \url{http://www.census.gov/genealogy/names/}.}
    to identify when noun phrases
    refer to people by name (useful for resolving \emph{he} and \emph{she}). 
    It also maps personal titles to genders when possible (e.g., to resolve \emph{her} to \emph{Mrs.}).
\end{itemizesquish}

\subsection{Full description}

To analyze a document, ARKref performs the following steps.
\begin{enumerate}
\item Parse all sentences to constituents, and recognize named entities and nominal supersenses.
\item Find all mentions.
\item For every mention, find its antecedent, if any:
\begin{enumerate}
\item Make immediate decisions for certain specific syntactic patterns.
\item For a pronominal mention, filter previous mentions by matching syntactic
type.
\item For nominal and proper mentions, filter previous mentions based on
matching surface features and semantic compatibility.
\item Among remaining filtered antecedent candidates, choose the candidate
with the smallest syntactic distance. If there are no candidates,
resolve to \texttt{NULL}.
\end{enumerate}
\item To partition mentions into entity clusters, take the transitive closure
of these antecedent selection decisions.
\end{enumerate}

\noindent
This approach depends completely on getting individual antecedent
selection decisions correct; it misses opportunities to use joint
information and constraints across the document, and it also can allow
a single bad decision to merge many non-coreferent mentions into the
same cluster.

%
%
%
\subsection{Subsystems}

We use the Stanford Parser,\footnote{The
\texttt{2008-10-26} version \citep{klein-03b}.}
and a supersense tagger described above.

\subsection{Mention identification}

For an unlabeled piece of text, we mark most NPs as mentions. Specifically,
the system takes all NPs that are the largest possible for their head
word, as defined by the Collins head rules \citep{Collins1999HeadRules}.
For example, if the children are a sequence of noun tokens, the head
will be the rightmost token; but if the subtree has a prepositional
attachment like (NP NP (PP IN NP)), then the head is the head of the
left NP. This prevents repetition of redundant noun phrases that are
embedded inside each other; for example,
\begin{quote}
$_{NP}${[} $_{NP}${[}the revised accounting{]} of $_{NP}${[}the
incident{]}{]}
\end{quote}
In this case, \emph{accounting} is the head of \emph{the revised accounting},
and it is also the head of \emph{the revised accounting of the incident}.
Both noun phrases are considered as belonging to the same mention;
the highest-level NP is used for syntactic pattern matching. The internal
noun phrase \emph{the incident} remains its own mention, since it
is the only and largest noun phrase whose head word is \emph{incident}.

This mention identification strategy finds pronouns, common nouns,
and named mentions. It was run on several reference texts from Wikipedia
and other sources, and seemed to perform reasonably well.

For evaluating on annotated ACE data, we follow previous work and
use the ACE data's definitions of mentions. This causes
conflicts when trying to reconcile annotators' definitions of phrases
with the Treebank-style parses and Collins head rules.

\subsection{Immediate match patterns}

ARKref includes a set of patterns for immediate 
matching of targets to potential antecedents.
If an antecedent candidate matches the target on one of these patterns, it is
immediately resolved.  (If not, the system progresses to the next step.)

Appositives are fairly easy to identify from the parse tree, and are
resolved immediately; for example, in the following cases, we start
from the right NP and find the left side is the immediate sibling
of an intervening comma token.
\begin{itemize}
\item {[}Lawrence Tribe{]}, the Harvard Law School {[}Professor{]} ...
\item {[}David Boies{]}, Gore 's chief trial {[}lawyer{]} ...
\end{itemize}
We also implemented a recognizer for role appositives, e.g. \emph{{[}Republican
candidate{]} {[}George Bush{]}}. Unlike \citet{haghighi-09}, we did not find this very
helpful.

The other useful immediate-match pattern is the predicate-nominative construction,
in which the subject and object of the sentence is mediated by a form
of the verb {}``to be.'' For example,
\begin{itemize}
\item {[}Lameu{]} was the first NHL {[}player{]} to become a team owner.
\item The {[}Gridiron Club{]} is an {[}organization{]} of 60 Washington
journalists.
\end{itemize}

\subsection{Pronoun resolution}

Pronominal mentions are identified through the parser's part-of-speech
analysis; specifically, PRP and PRP\$ nodes.

First, several syntactic patterns are checked for to reject (but never
immediately accept) certain candidates:
\begin{itemize}

\item The {}``I-within-I'' constraint: a pronoun cannot refer to a node
that dominates it.  An example from \citet{haghighi-09}:
\begin{itemize}
\item e.g. \emph{Walmart says Gitano, its top-selling brand, is underselling.
$\ \ \Rightarrow\ \ $ it $\neq$ Gitano}
\end{itemize}

\item A reflexive pronoun is required for a verb's object to corefer with the subject.
\begin{itemize}
\item e.g. \emph{The bank ruined it.$\ \ \Rightarrow\ \ $it $\neq$ bank}
\item e.g. \emph{The bank ruined itself}. $\ \ \Rightarrow\ \ $ \emph{itself
$=$ bank}
\end{itemize}

\item Subjects cannot refer to NPs in an adjunct phrase.
\begin{itemize}
\item e.g. \emph{To call John, he picked up the phone} $\ \ \Rightarrow\ \ $
\emph{he} $\neq$ \emph{John }
\item e.g. \emph{Because John likes cars, he bought a Ferrari.} $\ \ \Rightarrow\ \ $
$he$ = \emph{John}
\end{itemize}
\end{itemize}

\noindent
Next, syntactic type compatibility plays a major role in filtering
to allowable pronoun matches. The system identifies the following
types from the pronoun.

\begin{itemize}
\item Gender: Male, Female, Unknown (e.g. he/his vs. she/her vs. they/it)
\item Personhood: Pers, NotPers, Unknown (e.g. he/she vs. it/that)
\item Number: Singular, Plural (e.g. he/she/it vs. they/them/those)
\end{itemize}

Type information is inferred for antecedent candidates. For nominal
and proper mentions,

\begin{itemize}
\item \textbf{Gender}: ARKref identifies the gender of mentions as male or female (or unknown) by checking whether a word in the mention matched a common male or female first name in a list downloaded from the U.S. Census Bureau (or ``Mr.'' or ``Mrs.'').\footnote{\url{http://www.census.gov/genealogy/names/}} 

\item \textbf{Personhood}: To identify whether a mention refers to a person, ARKref checks whether the head word was labeled as a person by the supersense tagger, the gender was identified as male or female, or a common title such as ``Mr.'' appeared in the mention.\footnote{Code: \emph{Types.personhood()}}
\item \textbf{Number}: ARKref uses the parser's part-of-speech analysis to determine whether a mention is singular versus plural (i.e., whether the tag of the head word is NN or NNP versus NNS or NNPS).  For pronouns, a list of singular and plural pronouns is checked.
\end{itemize}

\noindent
This component of the syntactic analysis system is a major difference compared to
\citet{haghighi-09}.

\subsubsection{Use of syntactic type filtering}

Given that reliability of the identification of these various types
differs --- for example, number identification is quite reliable,
but personhood is harder and we sometimes give up, flagging as Unknown
--- we experimented with different rules for the strictness of matching.
For example, plural vs. singular is less definite for certain types
of entities like human organizations, which can be referred to as
both {}``they'' and {}``it.'' 

Gender information made little impact on the ACE development data,
which is newswire text, in which its is rare for pronouns of both
genders to be used in the same document. For example, the word {}``she''
appears in only 7 of 68 documents (while 44 contain {}``he''). 
Gender information actually slightly
hurts performance, even when used as a very lax constraint. However,
personhood and number matching was very useful. 

We only used syntactic type information for matching pronouns to other
mentions. We did not attempt to do this for nominal
mentions, since they seemed harder to reliably identify, despite our
usage of various lexical resources.

Enhancing syntactic type identification should be an avenue of future
work, since the current system, while it is limited relative to large-scale lexical
semantics efforts, is quite useful for performance---as illustrated in the ablation tests (Table~\ref{ablation}).

\subsection{Nominal and proper resolution}

Common nouns and names (a.k.a. nominal and proper mentions) are also
resolved by looking for an antecedent. Unlike pronouns, it is allowable
for these to have a \texttt{NULL} reference; for example, the first few mentions
in a document usually have no antecedent. It is arguable that the
antecedent selection approach, while reasonable for pronouns, doesn't
fit these cases as well.

In any case, ARKref includes two rules for this resolution, allowing
an exact match of head words or a substring match.\footnote{Code: \emph{Resolve.substringMatch().}}
The substring matching only occurs if both mention heads are tagged singular proper nouns (i.e., NNP),
are both at least 4 characters long, and begin with the same 4 characters 
(e.g., ``Japan'' and ``the Japanese'').

Note that these matching heuristics do result in some false positives (e.g. {}``Korean officials'' and {}``Iranian officials'').

Final selection is done through shortest path distance.


\subsection{Shortest path distance}

The above mechanisms yield a list of antecedent candidates. If there
are zero candidates, we resolve to \texttt{NULL}. If there are multiple candidates,
we choose the one that's closest by the syntactic path distance through
the parse tree. We allow crossing between sentences by linking all
sentence parses in a right-branching structure. (This can be thought
of as the simplest possible discourse structure.)

Path distance outperforms selection by simple surface distance. Consider
the following example:\footnote{This example is from \citet{clack-us-history}.}

\begin{quote}
While establishing a {[}refuge{]} for {[}Catholics{]},
who faced increasing {[}persecution{]} in {[}Anglican England{]},
the \textbf{{[}Calverts{]}} were also interested in creating profitable
\textbf{{[}estates{]}}. To this end, and to avoid {[}trouble{]}
with the {[}British government{]}, \textbf{{[}they{]}} also encouraged
{[}Protestant{]} {[}immigration{]}.
\end{quote}

\noindent
The mention \emph{they} has two
plural antecedent candidates, \emph{Calverts} and \emph{estates};
the latter is surface-closer, but since it is embedded in a predicate
clause, the first sentence's subject, \emph{Calverts}, is actually
syntactically closer. This is the right thing to do in this and other
similar examples. Path distance is better at capturing saliency.

\section{Evaluation}

The ARKref tool demonstrates competitive performance in 
experiments on subsets of the ACE Phase 2 coreference dataset,
using gold standard-defined mentions as in previous work.
We evaluate on the \texttt{ACE2004-ROTH-DEV} and \texttt{ACE2004-CULOTTA-TEST}
subsets (as named by \citet{haghighi-09}); the development set was used for development,
and the test set was not evaluated on until the writing of this paper.
We report two metrics: 

(1) Pairwise $F_1$, meaning the precision and recall
for recognizing mention-mention pairs as coreferent or not.
This is an intuitive metric, but some authors argue it
has an issue in quadratically
penalizing mistakes in larger clusters. It is defined as:
\[ 
  \text{TP}=\sum_S \sum_{i\neq j \in S} 1\{G(i)=G(j)\},\ 
  \text{FP}=\sum_S \sum_{i\neq j \in S} 1\{G(i)\neq G(j)\},\ 
  \text{FN}=\sum_G \sum_{i\neq j \in G} 1\{S(i)\neq S(j)\}
\]\[ P=\frac{\text{TP}}{\text{TP}+\text{FP}},\ 
  R=\frac{\text{TP}}{\text{TP}+\text{FN}},\ 
  F=\frac{2PR}{P+R} \]
Where
$G$ refers to a gold-standard entity and $S$ refers to a system-predicted entity,
where an entity is a set of mentions; and $G(i)$ and $S(i)$
refer to the gold and system entities that contains mention $i$.

(2) We also report the $B^3$ metric, where for one document,
\[P= \frac{1}{n} \sum_i \frac{|G(i) \cap S(i)|}{|S(i)|}, 
  R= \frac{1}{n} \sum_i \frac{|G(i)\cap S(i)|}{|G(i)|},
  F= \frac{2PR}{P+R} \]
where $i \in 1..n$ is each mention in the document.

We did all development with pairwise metrics, and report $B^3$ because it is the easiest to implement of the metrics in \cite{Lee2013Coref} that are used in more recent coreference research.
See \citet{haghighi-09} and \citet{Lee2013Coref} for more details on the dataset and
evaluation metrics.  It was not clear to us whether these previous works used micro- or macro-averaging.  We used micro-averaging for pairwise results (i.e., add false positive, false negative, and true positive pair counts across all documents, before computing P/R/F), and macro-averaging for $B^3$ (i.e., compute P/R for each document, then average them across documents).
Results are shown in Table~\ref{t:comparison}.

\begin{table}[h]
  \centering
\begin{tabular}{|l|ccc|ccc|}
  \hline 
  & \multicolumn{3}{c}{Pairwise} & \multicolumn{3}{c|}{B$^3$} \\
  & P & R & F & P & R & F \\
  \hline
  \multicolumn{7}{c}{ACE04-Roth-Dev (68 docs)} \\
  \hline
  ARKref       & 65.8 & 55.2 & 60.0 &   84.7 & 76.7 & 80.5 \\
  HK SYNCONSTR & 71.3 & 45.4 & 55.5 &   84.0 & 71.0 & 76.9 \\
  HK SEMCOMPAT & 68.2 & 51.2 & 58.5 &   81.8 & 74.3 & 77.9 \\
  \hline
  \multicolumn{7}{c}{ACE04-Culotta-Test (107 docs)} \\
  \hline
  ARKref       & 58.0 & 41.8 & 48.6 &   85.1 & 74.7 & 79.5 \\
  HK SYNCONSTR & 66.4 & 38.0 & 48.3 &   83.6 & 71.0 & 76.8 \\
  HK SEMCOMPAT & 57.5 & 57.6 & 57.5 &   79.6 & 78.5 & 79.0 \\
  Stanford Determ.~Coref  & & &             &   88.7 & 74.5 & 81.0 \\
  Bengston and Roth (2008) & & & & 88.3 & 74.5 & 80.8 \\
  Culotta et al.~(2007)    & & & & 86.7 & 73.2 & 79.3 \\
  \hline
\end{tabular}
\caption{Performance comparison.  ``HK'' = \citet{haghighi-09}.\label{t:comparison}}
\end{table}

%
%
%
%
%
%
%
%
%
%
%
%
%
%

\section{Other uses of ARKref}

Since the tool was first released in 2010, we are aware of several instances of its use in research.

ARKref was first used by \citet{Heilman2011Thesis}, 
who leveraged ARKref to increase the yield of a system 
for automatically generating questions from texts.

\cite{Lee2012Narrative} uses ARKref to analyze referring expressions in
narrative picture books.  On nine manually annotated narratives, ARKref and Stanford had nearly identical accuracy (0.54 and 0.55 B$^3$ F1).
\cite{Stern2011Entailment,Stern2012Entailment,Stern2012Search} use ARKref as a preprocessing tool for a
textual entailment recognition system.
\cite{kapp2013evenpers} uses it to support an event and person explorer tool.
\cite{crosthwaite2010review,xian2011benchmarking,subha2013ontology,subha2013quality} contain additional small scale evaluations of ARKref.

\section{Appendix}

We performed ablation and error analysis on an earlier version of ARKref.
The most major difference from the system described above is
that it did not use the supersense tagger, but instead performed personhood
detection on common nouns with wordlists derived from WordNet.

\subsection{Ablation analysis}

\noindent
Table~\ref{ablation} reports a series of ablation experiments on the development set.
This was performed with an earlier version of ARKref, thus the lower numbers for the ``Main system.''

\begin{table}[h]
\centering{}\begin{tabular}{|l|c|c|c|l|}
\hline 
 & P & R & F1 & \tabularnewline
\hline
\hline 
\textbf{Main system} & 64.1 & 48.1 & 55.0 & \tabularnewline
\hline 
Remove word lists & 63.6 & 47.9 & 54.6 & i.e. WordNet, U.S. Census\tabularnewline
\hline 
\textbf{Laxer} pronoun resolution &  &  &  & \tabularnewline
\hline 
Remove gender typecheck & 64.7 & 48.3 & 55.3 & slight improvement (!)\tabularnewline
\hline 
Remove person typecheck & 63.0 & 47.4 & 54.1 & \tabularnewline
\hline 
Remove number typecheck & 56.1 & 46.1 & 50.6 & \tabularnewline
\hline 
\textbf{Stricter} pronoun resolution &  &  &  & \tabularnewline
\hline 
Never resolve pronouns & 75.1 & 26.8 & 39.5 & \tabularnewline
\hline 
Never resolve 2nd person & 66.5 & 46.6 & 54.8 & \tabularnewline
\hline 
\textbf{Stricter Pro.-Pro}. &  &  &  & \tabularnewline
\hline 
Never match pro-pro & 67.3 & 41.8 & 51.5 & \tabularnewline
\hline 
Strict typechecking & 66.2 & 43.4 & 52.4 & \tabularnewline
\hline 
Check gram. number & 66.5 & 43.7 & 52.7 & \tabularnewline
\hline 
\hline 
\end{tabular}
\caption{Ablation analysis: Pairwise F1 performance on Bengston and Roth's ACE dev set \label{ablation}}
\end{table}

\subsection{Error analysis}

\noindent
We perform error analysis by inspecting the accuracy rates of individual
antecedent selection decisions; i.e., whether the chosen antecedent
from the candidate list is indeed coreferent with the mention. Note
that this accuracy rate has a non-trivial relationship with cluster-aware
metrics like pairwise F1 or $B^3$. For example, if a bad antecedent is
selected but the final cluster size is only those two mentions, that
hurts precision by only a single false positive. But if these two
mentions end up merging two different gold clusters, false positives
occur for every pair between the two gold clusters.
However, we suspect that antecedent-level accuracy rates 
may be 
indicative of overall accuracy.

\begin{table}

\begin{center}\begin{tabular}{|l|l|c|c|c|l|} 
\hline 
\multicolumn{2}{|c|}{Decision Type} & Corr. & Incorr. & Acc. & Notes \\
\hline
\multirow{3}{*}{Imm. Rules}
& Appositives & 105 & 23 & 82\% &\\
& Role Appos. & 5 & 0 & 100\% &\\
& Pred.-Nom. & 28 & 5 & 85\% &\\
\hline\hline
\multicolumn{6}{|l|}{Standard resolution pathways} \\
\hline
\multicolumn{2}{|l|}{Pronoun resolutions} & 460 & 235 & 66\% &\\
\multicolumn{2}{|l|}{Non-pronoun resolutions} & 1133 & 407 & 74\% &\\
\multicolumn{2}{|l|}{\texttt{NULL}} & 964 & 509 & * & Errors can be recovered later \\
\hline
\end{tabular}\end{center}
\caption{Breakdown of antecedent selection decisions\label{breakdown}\label{t:antecedent-selection}}
\end{table}

Table~\ref{t:antecedent-selection} breaks down the types of antecedent selection decisions the
system makes. The first thing to note is that the immediate syntactic
pattern matches are uncommon but relatively high accuracy. Inspecting
individual examples reveals a few changes could further improve precision.
Appositive errors include institutional affiliation and location specification
constructs. Perhaps typechecks could solve errors like the following
pairs, which currently get marked as coreferent:

\begin{itemize}
\item {}``{[}David Coler{]}, {[}VOA News{]}'' {}``{[}NPR news{]}, {[}Washington{]}{}``
\item {}``{[}Orange County{]}, {[}Calif.{]}'' {}``{[}Washington{]}, {[}D.C.{]}''
\end{itemize}
(The last example is arguably an error in the annotations; the correct
reading under most circumstances is as a single mention.)

It is surprising that role appositives are so rare. It is worth investigating
if there exist examples in the data that the current system is missing.

Predicate-nominative errors are interesting. A number of errors are
due to modal verbs being picked up by the syntactic rule. These should
be eliminated by forcing a stricter, smaller set of allowed verbs,
and perhaps handling negations. For example, the current system resolves
the following mention pairs as coreferent:
\begin{itemize}
\item {}``{[}I{]}'ll be that {[}president{]},'' he added...
\item {[}Koetter{]} may not have been Arizona State's top {[}choice{]}.
\end{itemize}
Though a few examples seem genuinely harder: {}``The Taliban are
predominantly Sunni Muslim...''

However, the bulk of possible improvements to the system are still
in other forms of pronoun and non-pronoun resolution. 
The type checking definitely helps (as shown in ablations in Table~\ref{ablation}),
but there is much room for improvement.

\begin{table}

\begin{center}\begin{tabular}{|c|c|c|l|}
\hline
Corr. & Incorr. & Acc. & Pronoun \\
\hline
4 & 18 & 18\% & your \\
17 & 23 & 42\% & you \\
9 & 8 & 53\% & our \\
22 & 19 & 54\% & their \\
39 & 31 & 56\% & they \\
11 & 8 & 58\% & them \\
39 & 22 & 64\% & i \\
9 & 4 & 69\% & him \\
7 & 3 & 70\% & my \\
23 & 10 & 70\% & we \\
106 & 34 & 76\% & he \\
30 & 9 & 77\% & it \\
36 & 10 & 78\% & its \\
67 & 15 & 82\% & his \\
13 & 1 & 93\% & she \\
\hline
\end{tabular}\end{center}

\caption{Antecedent selection breakdown for pronouns occuring at least 10 times\label{Flo:pro}}
\end{table}

There do not seem to be any especially easy types of pronouns. Even
first- and second-person pronouns, which at first glance seem odd
in a newswire corpus (for example, they often appear within quotations),
often get resolved correctly. A subset of pronoun
resolution accuracies is shown in Table \ref{Flo:pro}. 

Another odd case is pronoun-to-pronoun matches, which we didn't even
consider when building the system. It turns out many of these work
OK, even when matching between seemingly type-mismatches like {}``I''
resolving to {}``he'' --- e.g. in dialogue or quotations. We experimented
with adding more typechecking, and also adding grammatical number
typechecking (first vs. second vs. third person pronouns), but they
only gave precision gains at cost to recall (Table~\ref{ablation}).

As an example how dialogue and speaker shifts can be difficult, in
the following our system resolves {}``Ray Bourque'' to {}``he'':
\begin{itemize}
\item {}``We've always stuck together and we'll stick by Patrick,'' defenseman
{[}Ray Bourque{]} said. {}``We know {[}he{]} is a quality person
and a great family man.''
\end{itemize}

(A note on the {}``NULL'' row of Table~\ref{t:antecedent-selection}: a correct {}``NULL''
decision means the mention is actually a singleton in the gold annotations.
An incorrect {}``NULL'' decision is trickier to analyze. It specifically
means that among the previous mentions, there was a gold-coreferent
mention, but instead NULL was chosen as the antecedent. This doesn't
mean this mention will have a pair error with this should-have-been
antecedent; they could later be connected through a completely different
path if later mentions select them as antecedents. It's still important
to be aware of these errors, though, since some of them represent
recall errors for nominal mentions.)

\bibliographystyle{plainnat}
\bibliography{newbib,corefbib,otherpapers/heilman_dissertation}
\end{document}